\documentclass[10pt,twocolumn,letterpaper]{article}

\usepackage{wacv}
\usepackage{times}
\usepackage{epsfig}
\usepackage{graphicx}
\usepackage{amsmath}
\usepackage{amssymb}
\usepackage{color}
\usepackage[pagebackref=true,breaklinks=true,letterpaper=true,colorlinks,bookmarks=false]{hyperref}
\usepackage[table,xcdraw]{xcolor}
\newcommand*{\affaddr}[1]{#1} 
\newcommand*{\affmark}[1][*]{\textsuperscript{#1}}

\def\wacvPaperID{1317} 

\wacvfinalcopy 

\ifwacvfinal
\fi

\ifwacvfinal
\usepackage[breaklinks=true,bookmarks=false]{hyperref}
\else
\usepackage[pagebackref=true,breaklinks=true,colorlinks,bookmarks=false]{hyperref}
\fi

\begin{document}

\title{Pose-guided Generative Adversarial Net  for Novel View Action Synthesis}

\author{
Xianhang Li\affmark[1], 
Junhao Zhang \affmark[2],
Kunchang Li \affmark[3],
Shruti Vyas\affmark[1], 
and
Yogesh S Rawat \affmark[1]\\ 
\\
\affaddr{\affmark[1] CRCV, University of Central Florida} , 
\affaddr{\affmark[2] National University of Singapore} \\
\affaddr{\affmark[3] Shenzhen Institutes of Advanced Technology, Chinese Academy of Sciences}
}

\maketitle

\footnotetext[1]{The project page is on\\ \href{https://xhl-video.github.io/xianhangli/pas_gan.html}{https://xhl-video.github.io/xianhangli/pas\_gan.html}. } 
\footnotetext[2]{The authors email are: {xli421@ucsc.edu, junhao.zhang@u.nus.edu,\\  kc.li@siat.ac.cn, (shruti,yogesh)@crcv.ucf.edu }}


\begin{abstract}
   We focus on the problem of novel-view human action synthesis. 
  Given an action video, the goal is to generate the same action from an unseen viewpoint. 
  Naturally,
  novel view video synthesis is more challenging than image synthesis.
  It requires the synthesis of a sequence of realistic frames with temporal coherency.
  Besides, transferring different actions to a novel target view requires awareness of action category and viewpoint change simultaneously.
  To address these challenges
  we propose a novel framework named Pose-guided Action Separable Generative Adversarial Net (PAS-GAN), which utilizes pose to alleviate the difficulty of this task.
  First, we propose a \textbf{recurrent pose-transformation module} which transforms actions from the source view to the target view and generates novel view pose sequence in 2D coordinate space.
  Second, a well-transformed pose sequence enables us to separate the action and background in the target view.
  We employ a novel \textbf{local-global spatial transformation module} to effectively generate sequential video features in the target view using these action and background features.
  Finally,  the generated video features are used to synthesize human action with the help of a 3D decoder.
  Moreover, to focus on dynamic action in the video,
  we propose a novel \textbf{multi-scale action-separable loss} which further improves the video quality.
  We conduct extensive experiments on two large-scale multi-view human action datasets, NTU-RGBD and PKU-MMD, demonstrating the effectiveness of PAS-GAN which outperforms existing approaches. The codes and models will be available on \href{https://github.com/xhl-video/PAS-GAN}{https://github.com/xhl-video/PAS-GAN}.
\end{abstract}

\section{Introduction}

  Video generation is an interesting problem with a wide range of applications in robotics \cite{geometry,GQN}, data augmentation \cite{data_augumentation}, and augmented reality \cite{Ar}.  
  We have seen some recent efforts in video generation, which include stochastic video synthesis  \cite{MoCoGAN, TGAN, VAGN} and conditional video generation \cite{Mcnet, MonkeyNet, vid2vid}. 
  The unconditional stochastic approaches attempt to learn the distribution of data directly, whereas the conditional generation approaches utilize the guidance of some priors, such as semantic segmentation labels \cite{vid2vid}, pose of target video \cite{everybody} to simplify the problem. 

Motivated by the success of conditional video generation \cite{Mcnet, MonkeyNet, vid2vid}, we have recently seen some efforts \cite{view-lstm, vyas2020multi, Recurrent_Transformer, accvactionview, shiraz2021novel} which can generate the novel view video by utilizing priors in feature level. 
These methods can obtain the novel view of the motion and the appearance features through some differential operations. 
We first conclude that previous works generally transfer the action into target view with three steps:
Firstly, using an encoder to represent the action feature from source-view video in latent feature space, which can enhance the learning capacity of the network.
Secondly, transforming the viewpoint of this action features by the cooperation power of the priors and differential operations in the target view, significantly reducing the complexity of the problem. 
Finally,  
feeding the transformed features into a decoder, aiming at generating a set of sequential frames.
However,
although many elaborate designs proved to be effective, we still clearly found significant motion blur and complete stillness in the generated video.
Thus, there are three questions that are worth investigating: First, what kind of \textit{\textbf{action features}} do we prefer to translate into the target view more easily? Second, what type of \textit{\textbf{priors and corresponding operations}} are most beneficial for generating the motion and appearance details of the target view? Third, what \textit{\textbf{constraints}} can improve the balance of motion and static detail generation for the novel view video synthesis model?

Regarding the\textit{ \textbf{action features}}, 
in \cite{view-lstm, vyas2020multi, Recurrent_Transformer, accvactionview, shiraz2021novel}, the authors use 3D convolutional neural network to encode source video. 
However, the action features from RGB space may not be the best choice. 
Because the RGB modality contains redundant background information in addition to the rich foreground action, it is challenging to extract exclusive action dynamics without explicit supervision.
More importantly,
view transformation in such redundant feature space leads to inaccurate pixel rendering. 
It has been shown that pose information can model the action dynamics better than other modalities \cite{everybody, vid2vid}. 
Motivated by this, 
we explore the use of pose information from the source viewpoint to represent the action features. 
Specifically,
we use the pose sequence from the source video to ignore redundant background features. We propose a novel \textit{recurrent pose-transformation module} to map this source pose to the target viewpoint.

Second, there are various types of \textit{\textbf{priors}} used, such as RGB frame \cite{view-lstm, Recurrent_Transformer}, depth sequence \cite{view-lstm}, and pose sequence \cite{view-lstm} from the target viewpoint. We argue that these priors provide different types of information, e.g., RGB can provide rich appearance information and pose can provide rich action dynamics. Therefore, it is worth investigating their appropriate utilization instead of treating them in the same manner \cite{view-lstm}.
The existing approaches \cite{view-lstm, Recurrent_Transformer} perform a global transformation in latent space, which make it challenging to recover the fine action details for action prediction. 
Thus in our work,
we only take an image prior and use it to decouple pose and appearance features. The extracted pose is later used for transforming action dynamics and the transformed action dynamics is used along with the extracted appearance features to generate latent video features.
Specifically,
we propose a novel \textit{\textbf{recurrent local-global spatial transformation module}} which transforms latent features independently for each joints in the target pose and jointly learned with a \textit{global transformation}. The cooperation power of local and global transformation modules can generate comprehensive latent features for video decoder. 

Furthermore, we also investigate the \textit{\textbf{constraints}} used for video synthesis. 
We argue that the commonly-used reconstruction loss (mean squared error) is the main reason that causes the motion blur \cite{goodfellow2016deep, i2i2017image} because motion can not belong to the same Gaussian distribution. Although using the mean squared error (MSE) loss can lift the score of peak signal-to-noise ratio(PSNR), it can result in heavy motion blur for the visual quality. More discussion can be found in Section 3.4.
To address this issue, we propose to separate the action from the background in the generated video with the help of a transformed pose sequence. We utilize a multi-scale perceptual loss as the main objective to force the network to focus on the action and background details separately. In addition, we also use the adversarial loss to make the network learn the distribution of actions, so it can focus on action dynamics in addition to the static details.

In conclusion,
we propose PAS-GAN, a novel approach for novel-view action synthesis which is trained end-to-end jointly optimizing multiple objectives. We validate our approach on two large-scale multi-view human action datasets, NTU-RGBD \cite{NTURGBD} and PKU-MMD \cite{2017pku}. We demonstrate: (1) the effectiveness of the proposed framework in novel view action synthesis; (2) the benefits of local spatial transformation module; (3) the capability of multi-scale action-separable perpetual and adversarial loss formulation to generate better static details and coherent motion.

\vspace{-0.2cm}
\section{Related Works}
\begin{figure*}[t]
    \centering
    \includegraphics[width=0.85\textwidth]{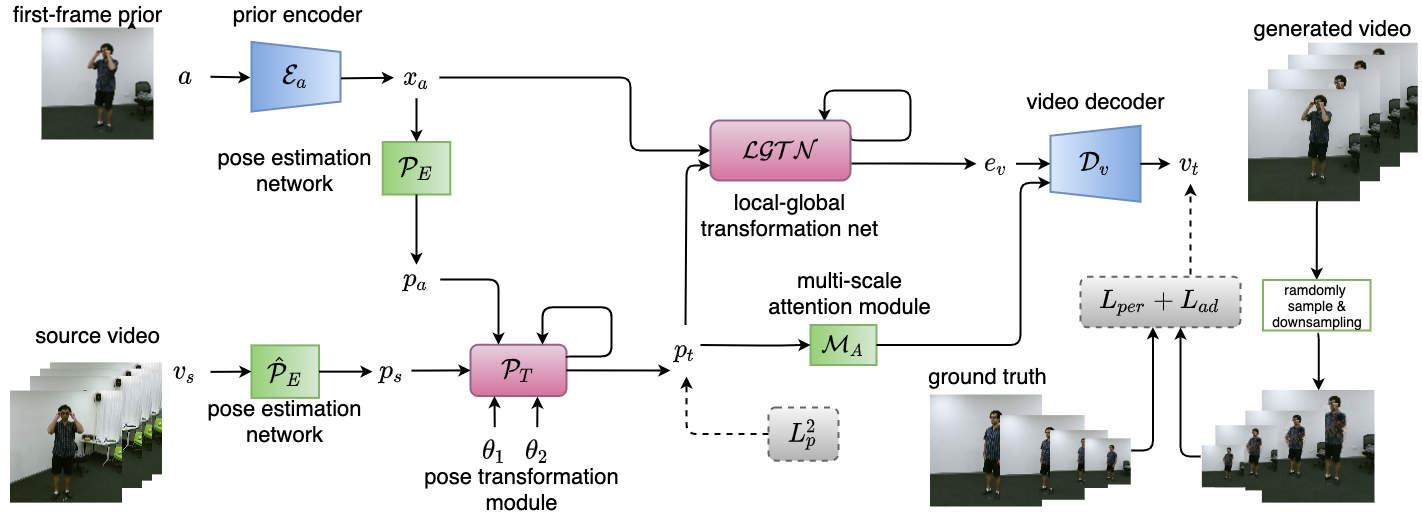}
    \caption{
    Overview of the proposed framework. Given an image prior $a$, which is the first frame of the target view video and an action video $v_s$ from the source viewpoint, the proposed network generates the video $v_t$ from the target viewpoint. Pose-transformation module $\mathcal{P}_T$ transforms the extracted pose information $p_s$ using $p_a$ to generate the target view pose sequence $p_t$. The global and local transformation module $\mathcal{LGTN}$ focus on coarse and fine action features, which uses appearance prior $x_a$ and $p_t$ to generate the latent features $e_v$. Different scales heat-map of $p_t$ which generated by multi-scale attention from $\mathcal{M}_A$ and $e_v$ are used to generate the target view video $v_t$. Finally, our model relies on the supervision of the multi-scale action-separable perceptual loss $L_{per}$ and adversarial loss $L_{ad}$.
    }
    \label{fig:framework}
    \vspace{-0.3cm}
\end{figure*}

\textbf{Conditional Video Generation}
Recent video generation approaches mainly focus on generative adversarial networks
(GANs) \cite{GANs,MoCoGAN, TGAN, VAGN, DVD-GAN}.
video prediction belongs to the conditional category, which \cite{Mcnet, MonkeyNet, vid2vid} can benefit from the priors.
In \cite{unsupervised}, the author proposed a deep video prediction model conditioned on a single image and an action class.  
In \cite{firstorder,zhang2021learning,luan2021pc}, the authors use a representation consisting of a set of learned key-points along with their local affine transformations to represent complex motion.
These methods utilize the key-points information to generate videos.
However,
the key-points and first frame are in the same view.
More recently,
methods in \cite{vid2vid, everybody, poseimage07, eccv2018pose}
 use pose as auxiliary information to help generate a more realistic video. However, our approach focuses on video generation from an unseen viewpoint. Unlike these methods that pose auxiliary information, we use pose to first separate appearance features and then recurrently generate action features. Second, we use the position information provided by pose to make the perceptual loss and adversarial loss more focused on the generation of action.
Moreover,
we need to estimate the key-points/pose for an unseen viewpoint, which is more challenging.

\textbf{Novel-View Synthesis.} 
  Most of the current work in novel view synthesis is focused on images where the focus is either on geometry-based \cite{Gemorty, PerspectiveNet} or learning-based methods \cite{viewindependet}. 
  with the help of a generative query network in \cite{GQN}, the authors utilize multiple views to render an image from unseen views. In \cite{viewindependet}, the authors proposed an encoder to extract view-independent features and a decoder hallucinate the image of a novel view.
  However, it is challenging to adopt the novel view image-synthesis methods into video synthesis due to several challenges.  The image synthesis approaches focus on the transformation of appearance and the lack of temporal coherence inevitably leads to a failure in modeling the motion transformation.  
  In \cite{Recurrent_Transformer, view-lstm, accvactionview},
  we have seen some efforts on novel view video generation, 
  however, due to multiple unknown parameters of the target-view, it remains unclear how much information on target view is required for video synthesis.
  Moreover,
  the issue of motion blur and static-frames is more evident in this task than conditional video generation.
\section{Method}

Given a source action video and an image prior from a target viewpoint, our goal is to generate a video of the same action from the target viewpoint. Formally, given an action video $v_s$ from a source viewpoint and an actor image $a$ from a novel target view (the first frame of the target view video), we aim to generate an action video $v_t$ from the target viewpoint. 
One of the key challenges is to transform relevant action dynamics from the source viewpoint to the target viewpoint. 
We propose a recurrent pose-transformation module $\mathcal{P}_T$ to overcome this challenge, which uses pose information $p_s$ and $p_a$ from the source video and target prior along with the corresponding viewpoint angels $\theta_1$ and $\theta_2$. The pose-transformation module $\mathcal{P}_T$ generates pose-sequence $p_t$ from the target view capturing action information.
The next challenge is to focus on fine action details which are hard to capture in latent representation. We propose a recurrent local and global transformation module $\mathcal{LGTN}$  which utilize the transformed pose-sequence $p_t$ to transform the appearance prior $x_a$ and generate latent features $e_v$ for target view action. The generated latent features $e_v$ are used to synthesize the target action video $v_t$ with a video decoder $\mathcal{D}_v$. 
The last challenge is how to design an effective loss to supervise the model to learn a more detailed motion and reduce the generation of blurry motion. To this end, we designed multi-scale action-separable loss.
It contains perceptual loss $L_{per}$ and adversarial loss $L_{ad}$ where the input of both loss is one randomly sampled frame from the generated video and ground truth frame at different scales. Moreover, the transformed pose is used to separate the action region which helps in generating details of motion and also preserve the appearance. An overview of the proposed approach is shown in Figure \ref{fig:framework}. 

\subsection{Recurrent Pose Transformation}

Given a pose-sequence $p_s$ from the source view and the prior-pose $p_a$ from the target viewpoint, the goal of pose-transformation module $\mathcal{P}_T$ is to estimate the pose-sequence $p_t$ from the target viewpoint. The $\mathcal{P}_T$ module has a recurrent structure that also utilizes the viewpoint information $\theta_1$ and $\theta_2$ corresponding to the source and the target viewpoints to perform this transformation. 
\begin{equation}
    p_t = \mathcal{P}_T(p_s, p_a, \theta_1, \theta_2),
\end{equation}
where $p_t$ is the transformed pose-sequence, $p_a$ is the prior pose, $p_s$ is the source pose sequence, $\theta_1$ is the source viewpoint, and $\theta_2$ is the target viewpoint. The prior pose $p_a$ is extracted from the prior image $a$ using a 2D convolution based model $\mathcal{E}_a$ \cite{VGG} and a pose estimation head $\mathcal{P}_E$. The appearance encoder $\mathcal{E}_a$ extracts latent appearance features $x_a$ and these latent features are passed to a fully connected network $\mathcal{P}_E$ for pose estimation. The source pose sequence $p_s$ corresponds to the source video $v_s$ and can be extracted using any existing pose estimation method $\mathcal{\hat{P}}_E$ \cite{HigerResolution}. In our experiments we utilize the pose information available with the training data. 

\begin{figure}[t]
    \centering
    \includegraphics[width=0.35\textwidth]{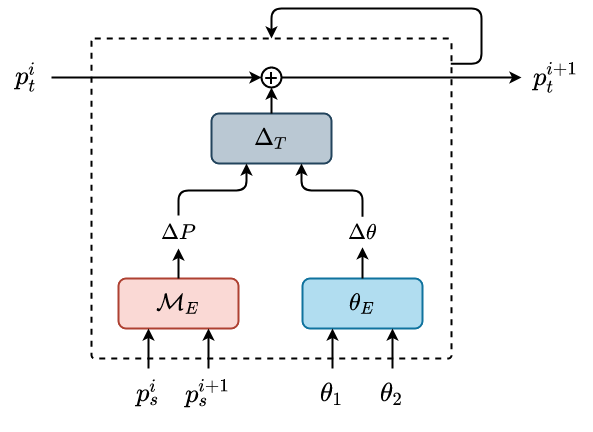}
    \caption{Overview of pose-transformation module.}
    \label{fig:pose_trans}
\end{figure}

A detailed overview of the pose-transformation module $\mathcal{P}_T$ is shown in Figure \ref{fig:pose_trans}. The recurrent structure of $\mathcal{P}_T$ consists of three main components; a motion estimator $\mathcal{M}_E$, change in viewpoint estimator $\mathcal{\theta}_E$, and a motion transformer $\Delta_T$. The motion estimator $\mathcal{M}_E$ aims at predicting pose changes between subsequent frames. It takes two subsequent poses $p_s^i$ and $p_s^{i+1}$ and predicts $\Delta P$, which represents the change in pose. 
\begin{equation}
\label{eq_twol}
    \Delta P = N(<N(p_s^i), N(p_s^{i+1})>)
\end{equation}
Here, $N$ represents a two-layered neural network and $<>$ indicates concatenation operation. Similarly, the $\mathcal{\theta}_E$ module estimates the change in viewpoint between the source and the target view. It estimates $\Delta \theta = \mathcal{\theta}_E(\theta_1, \theta_2)$ with the help of a neural network same as described in equation \ref{eq_twol}. 

The third component, motion transformer $\Delta_T$, takes the change in pose $\Delta P$ and change in viewpoint $\Delta \theta$ as input and transforms this motion to the target viewpoint. Finally, the transformed motion estimate is added to the hidden pose variable $p_t^i$ to generate the target pose for the next time-step. Here $p_t^i$ represents the pose in the previous time-step. The $\Delta_T$ module also uses the same architecture defined in equation \ref{eq_twol} and the hidden pose variable is initialized with the prior pose $p_a$. 

\subsection{Pose-guided the Motion Integration}

The estimated transformed pose $p_t$ is passed to a two-stream motion integration module which generates action features $e_v$ in latent space. This motion integration module consists of a local-global transformation network $\mathcal{LGTN}$.



\begin{figure}[t]
    \centering
    \includegraphics[width=0.45\textwidth]{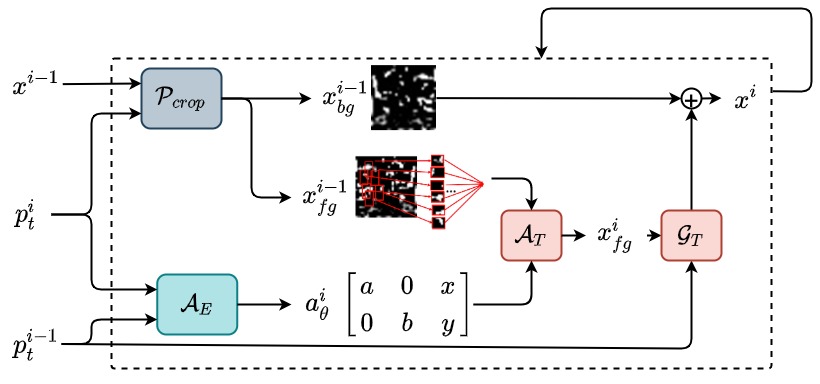}
    \caption{Overview of the local-global transformation network.}
    \label{fig:local_trans}
    \vspace{-0.3cm}
\end{figure}

 \textbf{Local-Global Transformation Network.} A global transformation to generate action features in latent space can be effective, but it can be challenging to recover fine action details with such global transformation. To address this issue, we propose a local-global transformation network $\mathcal{LGTN}$ which allows us to transform different key-regions in action independently. We utilize the joints present in the pose to define these key-regions. $\mathcal{LGTN}$ is based on Spatial Transformer Network (STN) \cite{STN} with three key differences. First, STN utilizes a common source to determine the transformation parameters and perform the transformations. $\mathcal{LGTN}$, on the other hand, utilizes the pose information to estimate the transformation parameters and the transformation is performed on the appearance features. Second, in STN a global transformation is performed on the source, whereas in $\mathcal{LGTN}$ we propose to perform a different transformation on different key-regions. Finally, STN does not have a notion of recurrent transformation, and $\mathcal{LGTN}$ has a recurrent structure that performs these transformations for a sequence of action features. A detailed overview of $\mathcal{LGTN}$ is shown in Figure \ref{fig:local_trans}. Moreover, after transforming the feature locally, we feed the foreground transformed feature into a global transformation based \cite{GRU}.

The local-global transformation network $\mathcal{LGTN}$ consists of four main components; a parameter estimator $\mathcal{A}_E$, key-region separator $\mathcal{P}_{crop}$, a transformation module $\mathcal{A}_T$ and a global transformation module $\mathcal{G}_T$ . The parameter estimator $\mathcal{A}_E$ predicts the transformation parameters $A_{\theta}$ with the help of a small convolution neural network which takes two subsequent pose heat-maps as input. It aims at learning the set of transformation parameters that are required to move from the initial pose $p_t^{i-1}$ to the next pose $p_t^i$. We utilize a 2D affine transformation where the estimated parameters can be represented as,
\begin{equation}
    a^i_{\theta} = 
\begin{bmatrix}
a^i & 0 & x^i \\
0 & b^i & y^i 
\end{bmatrix}
\end{equation}
where varying $a^i, b^i, x^i$, and $y^i$ allow for translation, rotation, and scaling. 

The key-region separator $\mathcal{P}_{crop}$ takes the pose from current time-step and extracts key-regions from the latent appearance features $x^{i-1}$ based on the joints. All the extracted key-regions represent foreground features $x_{fg}^{i-1}$ and the remaining features $x_{bg}^{i-1}$ are marked as background. The transformation module $\mathcal{A}_T$ applies the learned parameters $A_{\theta}$ on the foreground features $x_{fg}^{i-1}$ to generate the foreground appearance features $x^i_{fg}$ for the next time-step. This pointwise transformation is defined as,
\begin{equation}
\begin{bmatrix}
\alpha^{i} \\
\beta^{i}  
\end{bmatrix}
= a_{\theta}^i
\begin{bmatrix}
\alpha^{i-1} \\
\beta^{i-1}  \\
1
\end{bmatrix}
= 
\begin{bmatrix}
a^i & 0 & x^i \\
0 & b^i & y^i 
\end{bmatrix}
\begin{bmatrix}
\alpha^{i-1} \\
\beta^{i-1}  \\
1
\end{bmatrix}
\end{equation}
where $(\alpha^{i}, \beta^{i})$ are the target coordinates in the transformed foreground appearance feature map, $(\alpha^{i-1}, \beta^{i-1})$ are the source coordinates in the appearance feature map $x^{i-1}_{fg}$ from the previous time-step, and $a_{\theta}^i$ is the estimated affine transformation matrix. Finally the transformed foreground features $x^i_{fg}$ are aggregated and integrated back with the background appearance features $x_{bg}^{i-1}$ to compute video features $x_i$ for next time-step.
Then we feed the foreground feature $x^i_{fg}$ into the $\mathcal{G_{T}}$, based on the convolutional gated recurrent unit (Conv-GRU) \cite{GRU}, which takes current subsequent pose heat-maps as input to estimate motion and transform the $x^i_{fg}$.
This transformation is performed recurrently for each time-step. 
Finally,
we can obtain a sequence of features.
The final video features can be obtained $e_v = \mathcal{LGTN}(p_t, x_i)$, are for target view video generation.
    
\subsection{Video Decoder}

We utilize a 3D Conv based video decoder $\mathcal{D}_v$, which takes the generated video features $e_v$ and synthesize corresponding action video $v_t$. The pose information provides crucial details regarding the foreground region where the action occurs. Therefore we explore the use of estimated pose for attention mechanism in the video decoder. The transformed pose sequence $p_t$ is used to create a sequence of heatmaps $h_a$ with the help of a Gaussian kernel over joints. These pose heatmaps are integrated with the generated video features $e_v$ in the video decoder $\mathcal{D}_v$ which encourage the decoder to focus on foreground action regions. 


\textbf{Multi-scale Learning.} It can be challenging to recover fine action dynamics from a compressed latent representation; therefore we take a multi-scale approach while decoding the action video $v_t$. We extract the prior appearance features $x_a$ at different resolutions from the earlier layers of the prior encoder $\mathcal{E}_a$. Also, the motion integration using the local-global transformation network $\mathcal{LGTN}$ is performed at multiple scales using these multi-scale appearance features. The $\mathcal{LGTN}$ modules share weights for feature transformation across all scales. The pose heatmaps $h_a$ for soft attention are also computed at multiple scales using linear interpolation using $\mathcal{M}_A$. The transformed video features $e_v$ and pose heatmaps $h_a$ at multiple scales are then utilized by the video decoder at different stages to generate the video.

\subsection{Training Objective}

Overall, the proposed framework is jointly optimized for two different training objectives; transformed pose prediction $L^2_p$, and action video generation $L_{a}$. The  transformed pose prediction $L^2_p$ is optimized using a mean squared loss defined as,
\begin{equation}
    L^2_p = \frac{1}{NT}\sum\limits_{j=1}^{T}\sum\limits_{i=1}^{N} ((x_i^j - \hat{x}_i^j)^2 + (y_i^j - \hat{y}_i^j)^2)
\end{equation}
where, T is the total number of time-steps, N is total number of skeleton joints with coordinates $x_i^j$ and $y_i^j$ each. We experimented with multiple loss formulations for $L_a$ to improve the quality of video prediction. 
\begin{figure*}[t]
    \centering
    \includegraphics[width=0.9\textwidth]{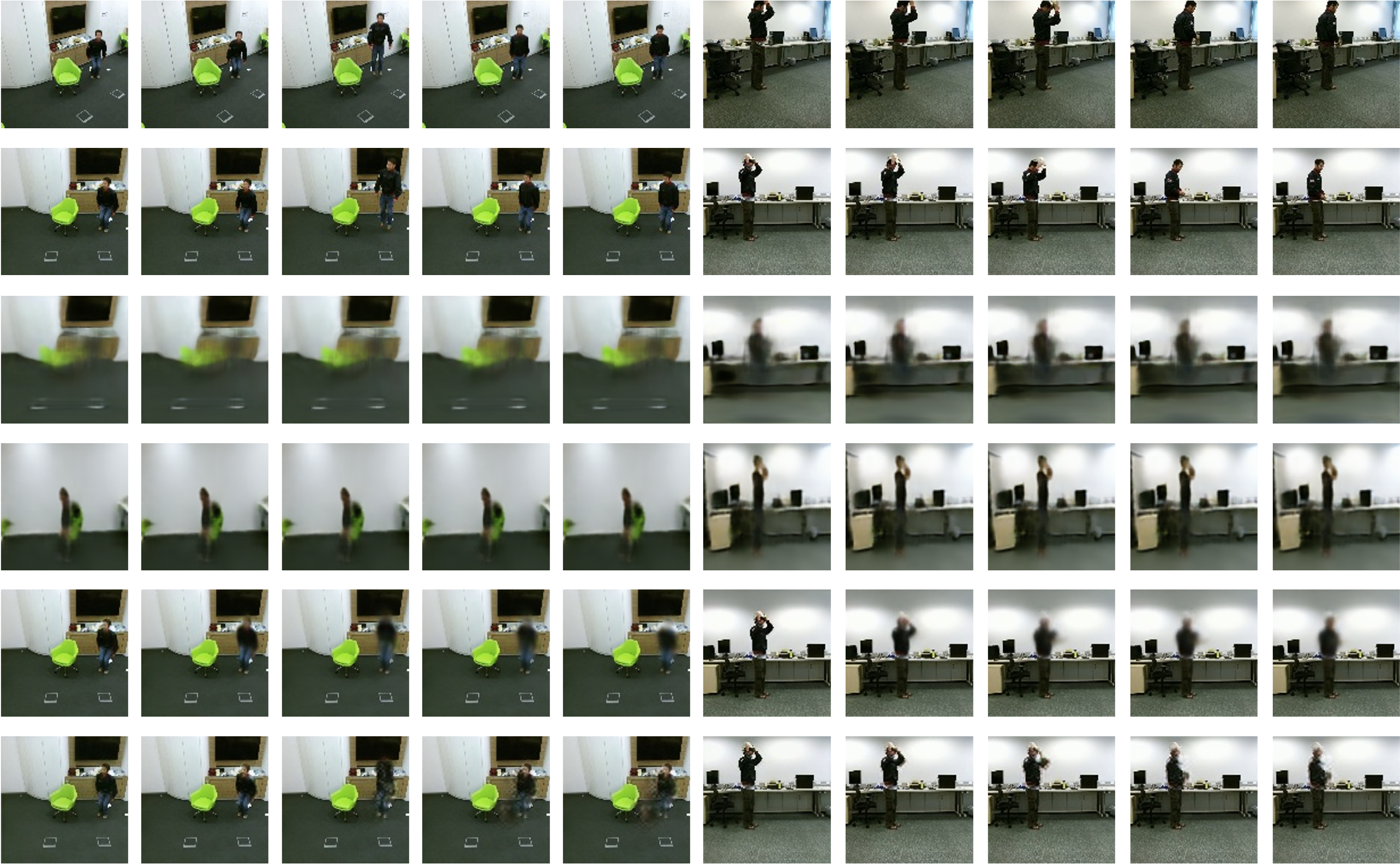}
    \caption{Comparison of the generated video frames using PAS-GAN with existing methods on NTU-RGBD \cite{NTURGBD} dataset. Row 1: source; Row
2: target; Row 3: VRNet \cite{TimeAware}; Row 4: VDNet \cite{view-lstm}; Row 5: RTNet \cite{Recurrent_Transformer}; Row 6: PAS-GAN (ours).
}
    \label{fig:ntusota1}
    \vspace{-0.3cm}
\end{figure*}

\textbf{Multi-scale Action Separable Loss.}
The multi-scale action separable loss contains one perceptual loss and adversarial loss of generated frames and generated actions at different scales.
First,
we use a perceptual loss $L_{per}$ to improve the visual quality of the generated video frames. \cite{perceptualloss}. We use a pre-trained vgg16-net \cite{VGG} to extract the frame-level features at different layers. We employed the ground truth frame $\hat{\mathbf{F}}$ and generated frames $\mathbf{F}$ as input.
\begin{equation}
\begin{split}
L_{per}=& \sum_{c=1}^{C}\left|V_{c}(\hat{\mathbf{F}})-V_{c}(\mathbf{F})\right|\\
& +\sum_{c=1}^{C}\left|V_{c}(\hat{\mathbf{{F}_{crop}}})-V_{c}(\mathbf{{F}_{crop}})\right|
\end{split}
\end{equation}
Where $V$ represents the pre-trained vgg16-net and $c$ is the current channel. We adopt the key-region separator $\mathcal{P}_{crop}$ to obtain ${F}_{crop}$.
To reduce the computational cost,
the perceptual loss $L_{per}$ is calculated by only randomly sampling one of the generated video frames each step. 
Finally, 
we adopt a downsampling layer for the input $F$ to employ the multi-scale $F$ at 4 scales $[1, 0.5, 0.25, 0.125]$. Thus we have 16 items in total in $L_{per}$. In addition to this, we also explore the use of adversarial loss \cite{GANs} to make the generated videos realistic. We use vgg16-net based discriminator with four scale inputs and adopt the $\mathcal{P}_{crop}$ to obtain actions as input similarly. 
The final training objective for our framework is,
\begin{equation}
    L =\lambda_1L^2_p  + \lambda_2L_{per} + \lambda_3L_{ad}
\end{equation}
where $\lambda_1, \lambda_2$ and $\lambda_3$ are weights which are estimated experimentally. 
Motivated by some excellent super-resolution works \cite{iccv2017srgan, eccv2018esrgan},
we use a MSE loss pre-trained network to initialize our model parameters.

\section{Experiments}

\begin{table*}[t!]
\centering
\resizebox{0.98\textwidth}{!}{
\begin{tabular}{c|c|c|c|c|c|c|c|c}
\hline\hline
                                          &\multicolumn{5}{c}{\textbf{Pair-view SSIM Score}}                                              &   \\ 
\hline
\textbf{Model}                   &$v1 \to v2$     &$v1 \to v3$     &$v2 \to v1$      &$v2 \to v3$      &$v3 \to v1$    &$v3 \to v2$       &\textbf{Average SSIM} &\textbf{Average PSNR}\\
\hline
VDG\cite{viewdisentanglement}    &$.502 \pm .058$ &$.543 \pm .068$ &$.584 \pm .060$ &$.563 \pm .062$ &$.611 \pm .077$  &$.522 \pm .063$   &$.554 \pm .075$   &- \\ 
\hline
PG$^2$\cite{poseimage07}         &$.499 \pm .071$ &$.561 \pm .060$ &$.600 \pm .064$ &$.557 \pm .071$ &$.598 \pm .075$  &$.543 \pm .055$   &$.560 \pm .076$   &- \\ 
\hline
VRNet\cite{TimeAware}            &- &- &- &- &-  &-   &$.68$   &$19.8$ \\ 
\hline
ResNet\cite{view-lstm}           &$.705 \pm .115$ &$.735 \pm .095$ &$.717 \pm .130$ &$.690 \pm .122$ &$.734 \pm .127$  &$.669 \pm .150$   &$.708 \pm .127$   &- \\ 
\hline 
VDNet\cite{view-lstm}            &$.789 \pm .076$ &$.791 \pm .069$ &$.800 \pm .076$ &$.765 \pm .079$ &$.797 \pm .067$  &$.756 \pm .089$   &$.783 \pm .078$   &- \\
\hline
RTNet\cite{Recurrent_Transformer}&$.974 \pm .021$ &$.975 \pm .021$ &$.975 \pm .019$ &$.971 \pm .021$ &$.974 \pm .017$  &$.971 \pm .022$   &$.973 \pm .020$   &$27.5 \pm 2.70$\\
\hline
PAS-GAN(ours)                             &${.977} \pm {.007}$ &${.975} \pm {.006}$ &${.974} \pm {.008}$ &${.975 }\pm {.006}$ &${.978} \pm {.007}$  &${.977} \pm {.006}$   &$\textbf{.976} \pm \textbf{.006}$   &$\textbf{28.07} \pm\textbf{ 1.39}$\\
\hline
\hline
\end{tabular}
}
\caption{Comparison of SSIM and average PSNR scores with existing methods. We report scores from all the combinations of three testing views. The scores for VDG \cite{viewdisentanglement} and PG2 \cite{poseimage07} are shown as reported by the authors of VDNet\cite{view-lstm}.}
\label{tab:compare_sota}
\vspace{-0.3cm}
\end{table*}

We aim to demonstrate the effectiveness of the proposed approach in novel view video synthesis and highlight the benefit of its components including, recurrent pose-transformation module, local-global transformation module, and multi-scale action separable loss. 

\subsection{Experiment Setup}
\textbf{Dataset.}
We conduct our experiments on NTU-RGBD \cite{NTURGBD} and PKU-MMD \cite{2017pku}, 
which are both large scale multi-view action datasets.
The NTU-RGBD contains over 56,000 videos and 60 action classes in total.
The videos are recorded using three different cameras and there are a total of 80 different viewpoints.
The PKU-MMD contains over 3500 action clips and 51 action classes which are also recorded by three cameras.
We use the 2D RGB pose, which indicates key points in the original RGB resolution and the viewpoint angels.
The pose information can also be extracted using some advanced pose estimation methods \cite{HRnet, Hourglass} for each frame.

\textbf{Implementation Details.}
We train our network in an end-to-end manner optimizing multiple objectives.
We use $\lambda_1 = \lambda_2 =1, \lambda_3 =1e-4$ in our training objective.
The network is trained using Adam optimizer with a learning rate of 1e-4 with a batch-size 12. 
For NTU-RGBD, we use a resolution of $112 \times 112$ in our experiments with 8 frames.
For PKU-MMD, we use $224 \times 224$. 
We extract four-scales feature with spatial resolution $112\times112 \& (224 \times 224)$, $56 \times 56 \& (112 \times 112)$, $28 \times 28 \& (56 \times 56)$, $14 \times 14 \& (28 \times 28)$ respectively from a modified vgg16-net \cite{VGG}.
We implement our framework on PyTorch \cite{pytorch} and use a Tesla V100.

\begin{figure}[t!]
    \centering
    \includegraphics[width=0.45\textwidth]{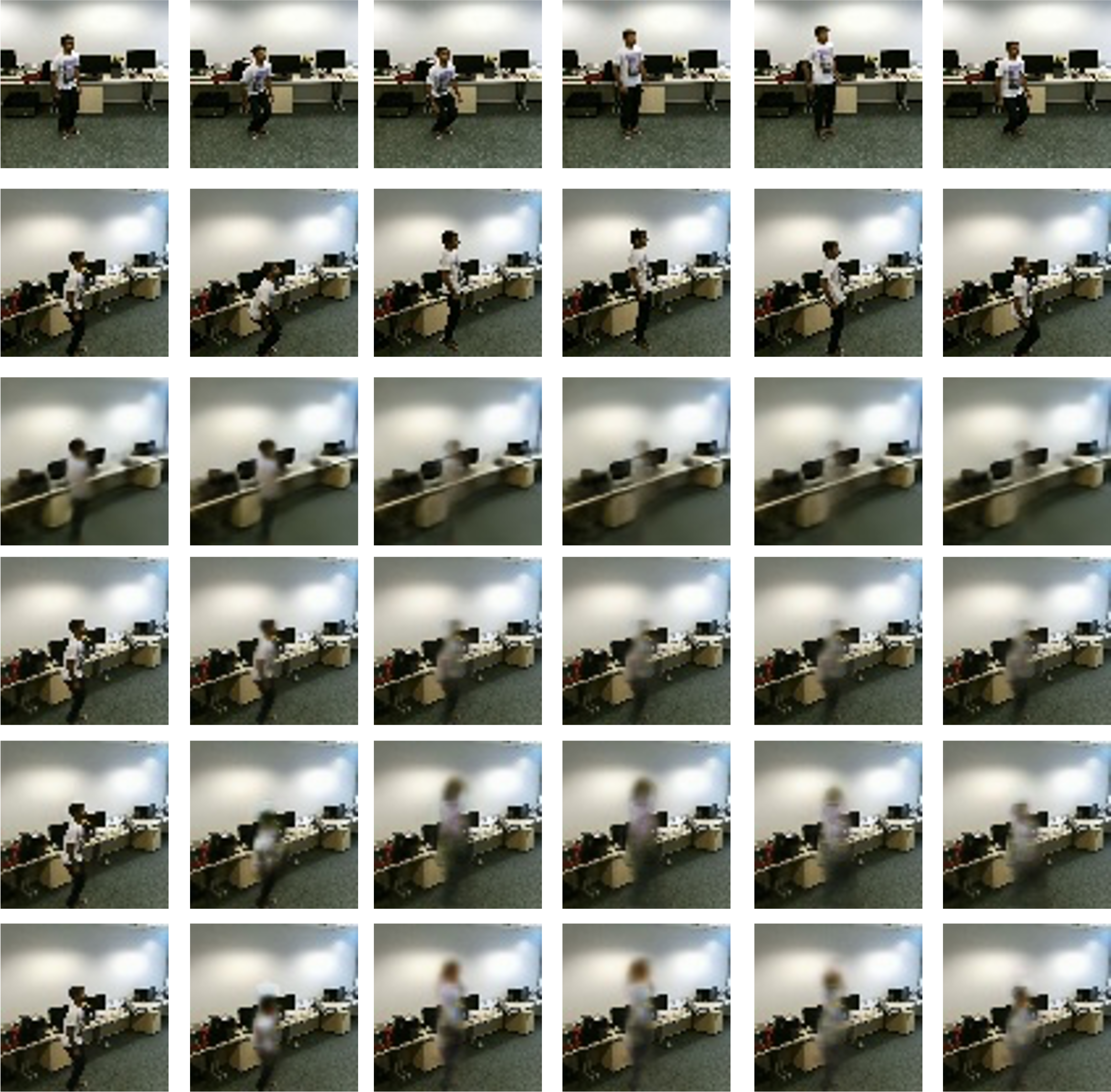}
    \caption{Ablation to evaluate different components of PAS-GAN. Row 1: source; Row
2: target; Row 3: BasicNet; Row 4: w/ multi-scale learning; Row 5: w/ recurrent pose transformation module; Row 6: w/ local-global transformation module. \textbf{It is important to note that row 6 is not generated using the full model}.}
    \label{fig:ablation1}
    \vspace{-0.3cm}
\end{figure}

\begin{figure}[t]
    \centering
    \includegraphics[width=0.45\textwidth]{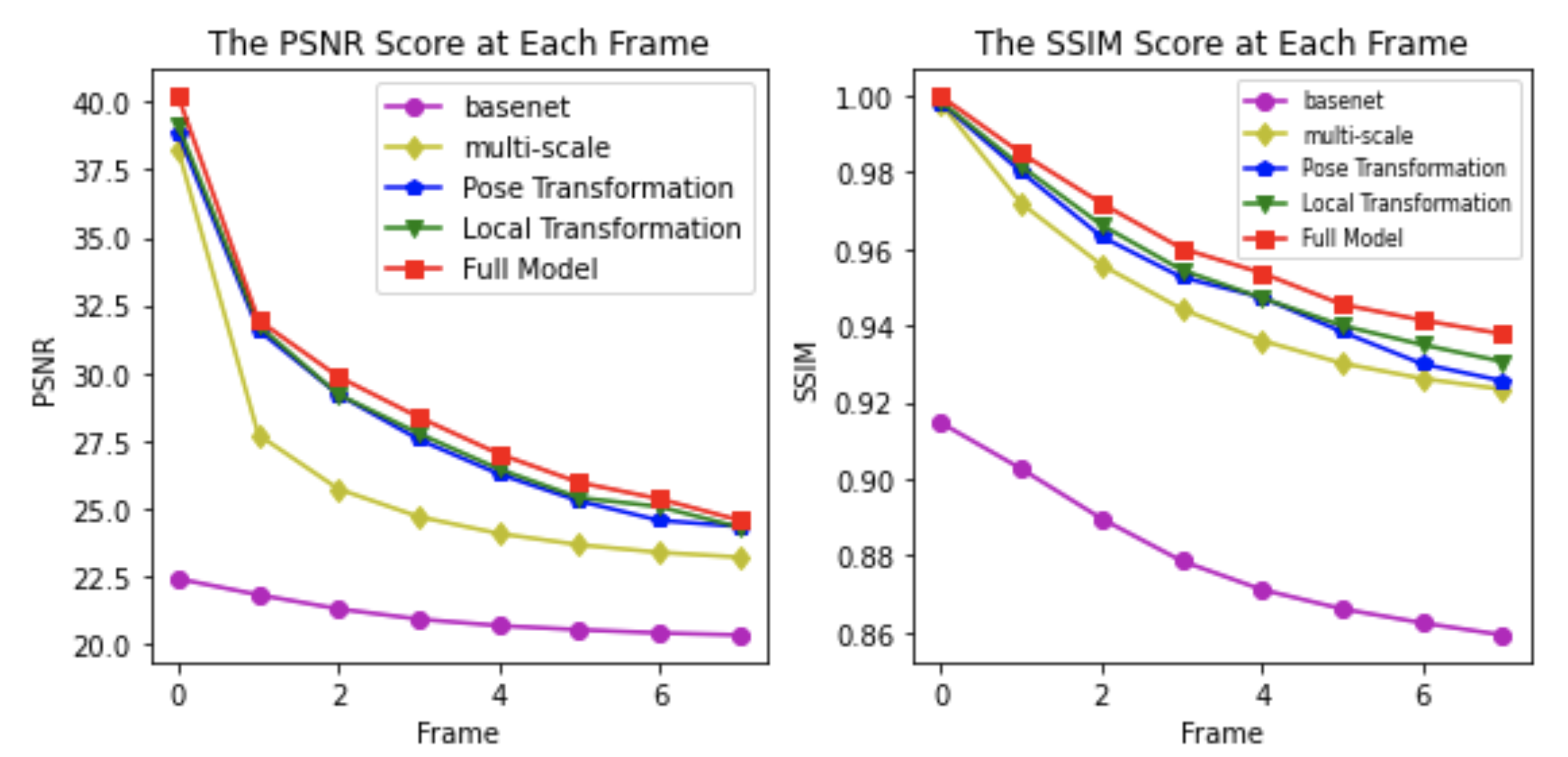}
    \caption{Ablation to compare different components using frame-level PSNR and SSIM scores.}
    \label{fig:ablation3}
    \vspace{-0.3cm}
\end{figure}

\begin{figure}[t]
    \centering
    \includegraphics[width=0.43\textwidth]{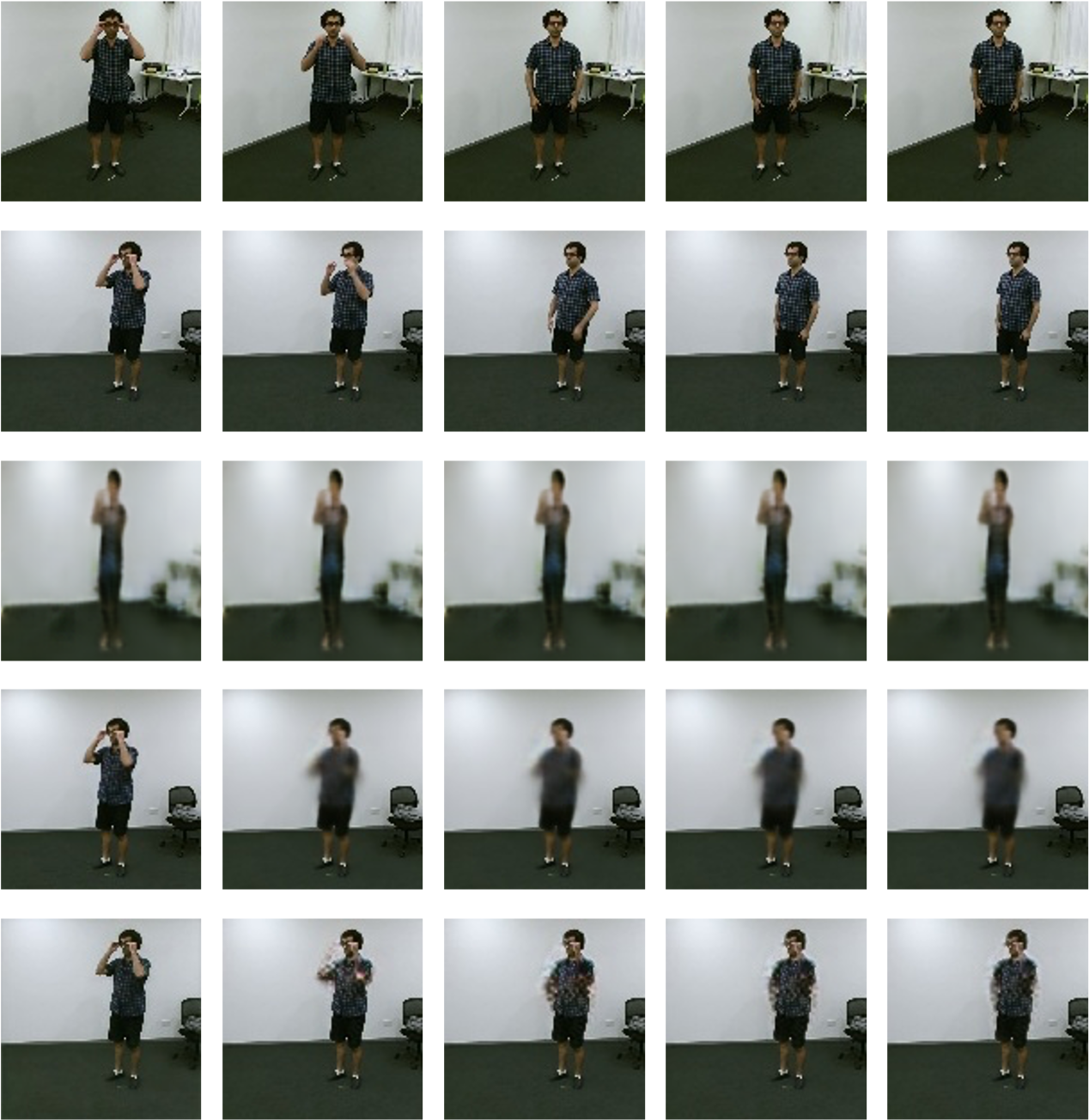}
    \caption{Generated novel view videos using existing methods highlighting motion blur and static video frames. Row 1: Source Video; Row 2: Target Video; Row 3: VDNet \cite{view-lstm}; Row 4: RTNet \cite{Recurrent_Transformer}; Row 5: PAS-GAN (Proposed method). Compared with previous state-of-the-art methods, our model can alleviate the motion blur and maintain the temporal coherence to some extent.}
    \label{fig:motivation}
    \vspace{-0.3cm}
\end{figure}

\begin{table}[t!]
\centering
\resizebox{0.45\textwidth}{!}{
\begin{tabular}{c|c|c|c|c|c}
\hline\hline
\textbf{Model}                   &MSE $\downarrow$      &PSNR $\uparrow$       &SSIM $\uparrow$                &FVD  $\downarrow$  &Pose MSE  $\downarrow$  \\
\hline 

VRNet\cite{TimeAware}            &$.00281$              &$25.54 \pm 1.45$         &$.923\pm .010$                     &$26.30 \pm .020$  &-\\ 
\hline
RTNet\cite{Recurrent_Transformer}&$.00082$             &$31.56 \pm 2.46$        &$.974 \pm .021$       &$3.90 \pm .12$    &-\\
\hline
BasicNet                         &$.00406$       &$24.05 \pm 1.03$         &$.938 \pm .016$             &$16.64 \pm .022$ &0.032\\
\hline 
PAS-GAN(ours)                             &$\textbf{.00058}$   &$\textbf{32.58}\pm \textbf{1.27}$  &$\textbf{.984} \pm \textbf{.005}$   &$\textbf{3.61} \pm \textbf{.10}$  &$\textbf{0.009}$\\
\hline
\hline
\end{tabular}
}
\caption{Key-region based MSE, PSNR, SSIM, \& FVD scores along with MSE score for pose estimation. 
}
\label{tab:forground_csota}
\vspace{-0.3cm}
\end{table}

\begin{figure*}[t!]
    \centering
    \includegraphics[width=0.9\textwidth]{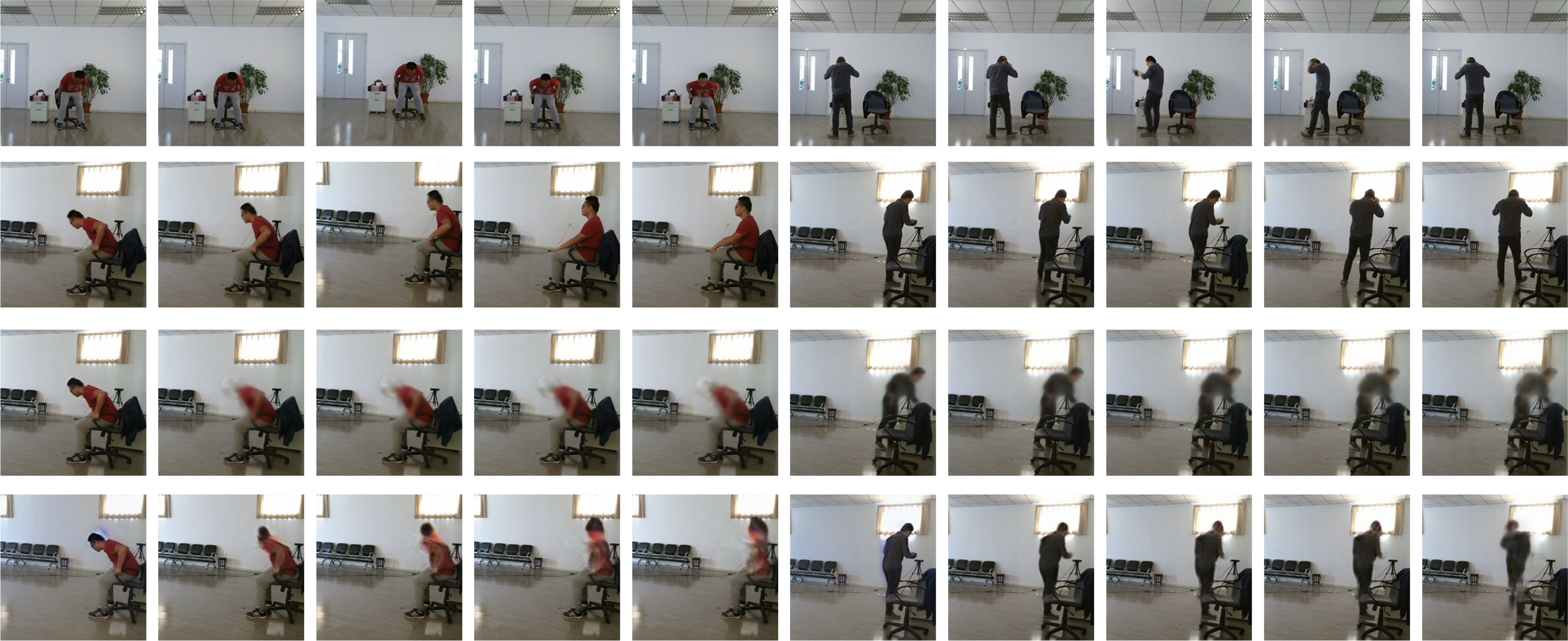}
    \caption{Comparison of the generated video frames using PAS-GAN with existing methods on PKU-MMD \cite{2017pku}. Row 1: source; Row
2: target; Row 3: RTNet \cite{Recurrent_Transformer}; Row 4: PAS-GAN (proposed method).
}
    \label{fig:pkusota1}
    \vspace{-0.3cm}
\end{figure*}
\textbf{Evaluation Metrics.}
To evaluate the performance of the proposed methods,
we follow \cite{view-lstm, Recurrent_Transformer} and use Structural Similarity (SSIM) \cite{SSIM} and Peak Signal to Noise Ratio (PSNR) for per-frame quantitative results. 
In addition,
we adopt the  Fr\'echet Video Distance \cite{FVD} for evaluating the quality of generated videos.
Moreover,
we also compute MSE, SSIM and PSNR scores only on the foreground region to focus more on the generated action. We utilize the pose information to crop out the foreground region in the generated and ground-truth videos. 

\textbf{Baseline}
To obtain our baseline model (BasicNet),
we use a modified vgg16-net \cite{VGG} to encode the image prior and a 3D convolution-based video decoder to generate the video.
The input is the image prior and we expand the image encodings to add temporal dimension before passing it to the video decoder.
Based on this BasicNet,
we use this as a basic model in our approach and evaluate the effectiveness of various components on top of this.
Another important baseline method is VD-Net \cite{view-lstm}.
Instead of using a single image as prior,
view-lstm utilizes the sequence of depth heatmap and skeleton from the target view,
which already contains sufficient motion information from the target view. 
On the other hand, our method focuses on transforming the motion from the source view to the target view.
We compare the proposed method with these baselines and other state-of-the-art methods, including \cite{Recurrent_Transformer} both qualitatively as well as quantitatively.

\begin{table}[t!]
\centering
\resizebox{0.45\textwidth}{!}{
\begin{tabular}{c|c|c|c|c}
\hline\hline
\textbf{Model}                   &MSE $\downarrow$     &PSNR $\uparrow$    &SSIM $\uparrow$     &FVD  $\downarrow$   \\
\hline
BasicNet                         &$.00394 $           &$24.38 \pm 1.63$     &$.924 \pm .030$     &$17.87 \pm .67$ \\ 
\hline
w/ Multi-Scale                   &$.00272 $           &$26.32 \pm 2.39$     &$.950 \pm .029$     &$5.94 \pm .084$ \\ 
\hline
w/ Recurrent Pose Transformation                   &$.00226 $           &$27.23 \pm 2.58$     &$.958 \pm .028$     &$4.83 \pm .127$ \\ 
\hline
w/ local-global Transformation Network             &$.00182$             &$27.85 \pm 1.89$     &$.968 \pm .016$     &$4.11 \pm .067$ \\
\hline
\hline
\end{tabular}
}
\caption{Key-region based evaluation of components in PAS-GAN.}
\label{tab:key-region-table_model_design}
\vspace{-0.3cm}
\end{table}

\begin{table}[t!]
\centering
\resizebox{0.45\textwidth}{!}{
\begin{tabular}{c|c|c|c|c}
\hline\hline
\textbf{Model}                         &MSE $\downarrow$     &PSNR $\uparrow$    &SSIM $\uparrow$     &FVD  $\downarrow$   \\
\hline
w/  MSE Only                             &$.00182$             &$27.85 \pm 1.89$  &$.968 \pm .016$     &$4.11 \pm .067$ \\
\hline
w/ Perceptual Loss\cite{perceptualloss}Only &$.00197$             &$27.43 \pm 1.75$ &$.967 \pm .013$      &$3.94 \pm .090$ \\ 
\hline
w/ Multi-Scale Action Separable Loss                &$.00177$            &$28.09 \pm 1.87$ &$.970 \pm .015$      &$3.83 \pm .067$ \\ 
\hline 
\hline
\end{tabular}
}
\caption{Key-region based evaluation of different loss functions.}
\label{tab:table_loss_comparison}
\vspace{-0.3cm}
\end{table}
\vspace{-0.3cm}
\subsection{Evaluation}

We perform an extensive evaluation of PAS-GAN on NTU-RGBD and PKU-MMD datasets. The quantitative evaluation on NTU-RGBD dataset is shown in Table \ref{tab:compare_sota} and \ref{tab:forground_csota}. In Table \ref{tab:compare_sota}, we have shown pair-wise SSIM scores for each pair of viewpoints. We observe that PAS-GAN consistently performs well on all pairs of viewpoints independent of the change in viewpoint. In Table \ref{tab:forground_csota}, we have shown scores for the action region in the video and we can observe that scores are comparable (slightly better) compared with the evaluation on the full video. We also show the generated video frames for qualitative evaluation on NTU-RGBD and PKU-MMD datasets. The generated frames are shown in Figure \ref{fig:ntusota1} and \ref{fig:pkusota1}. We observe that the generated frames have visible action dynamics capturing the target action. 

\textbf{Quantitative Comparison.}
We first perform a quantitative comparison of PAS-GAN with existing methods using SSIM, PSNR, and FVD metrics on the NTU-RGBD dataset. The comparison is shown in Table \ref{tab:compare_sota} and \ref{tab:forground_csota}. In Table \ref{tab:compare_sota}, we compare the SSIM and PSNR scores on the full generated video. We observe that PAS-GAN outperforms all the existing methods in both the evaluation metrics. Table \ref{tab:forground_csota} shows the evaluation only on the activity region. We can observe that the proposed method provides a significant improvement over the baseline model BasicNet and also outperforming the other existing methods.


\textbf{Qualitative Comparison.}
We also perform qualitative comparison on NTU-RGBD and PKU-MMD. The generated video frames are shown in Figure \ref{fig:ntusota1}, in Figure \ref{fig:motivation}, and in Figure \ref{fig:pkusota1}. 
Although the quantitative scores of PAS-GAN and previous SOTA RTNet \cite{Recurrent_Transformer} is close, we can see a significant improvement our PAS-GAN made in terms of the visual quality.
The action dynamics can be visible in the generated frames and it is much better when compared with the other models on the same dataset. 

\vspace{-0.3cm}
\subsection{Ablations}
We conduct several ablation experiments from two different perspectives. First is to validate the effectiveness of each component in PAS-GAN and second is to explore the impact of different loss functions.

\textbf{Effectiveness of Components.}
To study the impact of different components in PAS-GAN, we sequentially add proposed components in the BasicNet model. The comparison is shown in Table \ref{tab:key-region-table_model_design} and in Figure \ref{fig:ablation1} where we observe that the largest improvement results from the multi-scale learning. 
We also observe a consistent performance gain as each module is added to the network.
However,
the performance gain diminishes eventually and we argue that this is due the use of a smaller resolution ($56 \times 56$) in our ablations, which limits the performance to some extent.
Moreover,
we plot the per-frame PSNR and SSIM respectively in Figure \ref{fig:ablation1}. 
It clearly shows that different modules contributes distinctively.
Beside,
due to the use of an image prior,
the results are at the peak for the first frame.

\textbf{Influence of Loss Functions.}
PAS-GAN is trained using multiple objectives. To study the impact of these loss functions, we perform some ablations which are shown in Table \ref{tab:table_loss_comparison}. 
First surprising finding is that, although using only perceptual results in worse scores than MSE, there is a huge difference in visual quality between them as shown in Figure \ref{fig:ntusota1}.
Next, we observe that
the proposed multi-scale action-separable loss improves the performance further.
It is due to the use of multiple scales and separation of action and background. They significantly help the model to focus more on  fine motion and appearance details.




\subsection{Discussion}
We would like to explore in more depth the similarities and differences between our approach and RTNet \cite{Recurrent_Transformer}. Our problem definition is the same as theirs: given the first frame of the target view, the goal is to transform the source video to the target viewpoint. The first difference is that we do not use the source video to represent the action features. Instead, we use the 2D pose obtained from each frame of the source video. Transforming the viewpoint in 2D pose coordinate space greatly reduces our transformation difficulty as non-action features are automatically ignored. Secondly, RTNet only focuses on the global information when generating the target view features, which leads to a lack of local information which leads to motion blur. Our local-global transformation module uses the 2D pose coordinate information to automatically focus on the local information and use a global transformation module to integrate further. Finally, we use the multi-scale perceptual loss as the primary loss instead of MSE, and separate the action with the transformed pose coordinates for its supervision. This enables our approach to address the issue of motion blur and generate frames with coherent motion.

\vspace{-0.3cm}
\section{Conclusion}
We present PAS-GAN for novel view video synthesis where we explore the use of pose to solve this problem. Our method utilizes a pose transformation module to capture action dynamics from a source viewpoint. 
We show how a local-global feature transformation can be used jointly to learn effective action dynamics. We also propose a novel video specific multi-scale action-separable perceptual and adversarial loss formulation to improve the quality of generated videos. Extensive evaluation on two large-scale action datasets demonstrates the effectiveness of the proposed approach and the benefits of its various components.

{\small
\bibliographystyle{ieee_fullname}

}

\end{document}